\documentclass[lettersize,journal]{IEEEtran}

\usepackage{amsmath,amsfonts}
\usepackage{algorithmic}
\usepackage{algorithm}
\usepackage{array}
\usepackage[caption=false,font=normalsize,labelfont=sf,textfont=sf]{subfig}
\usepackage{textcomp}
\usepackage{stfloats}
\usepackage{url}
\usepackage{verbatim}
\usepackage{graphicx}
\usepackage{cite}
\usepackage{booktabs}
\usepackage{microtype}
\usepackage{multirow}
\usepackage[table]{xcolor}
\usepackage[colorlinks=true,linkcolor=blue,citecolor=blue,urlcolor=blue]{hyperref}
\begin{document}

\title{Spectral Invariant Fusion and Refinement for Domain Generalized Semantic Segmentation}

\author{Yansong Bai, Yunzhi Zhuge, Xinzhuo Yu, Lu Zhang, Jin Zhan, Pingping Zhang, Huchuan Lu,~\IEEEmembership{Fellow~IEEE}
\thanks{This work was supported in part by the National Natural Science Foundation of China under Grants 62406053 and 62441231, and in part by the Liaoning Provincial Doctoral Scientific Research Start-up Fund Project under Grant 2025-BS-0080, and in part by the Fundamental Research Funds for the Central Universities under Grant DUT25LAB128.}
\thanks{Yansong Bai, Yunzhi Zhuge, Lu Zhang, Pingping Zhang and Huchuan Lu are with the School of Information and Communication Engineering, Dalian University of Technology, Dalian 116081, China. (e-mail: bys@mail.dlut.edu.cn; zgyz@dlut.edu.cn; zhangluu@dlut.edu.cn; lhchuan@dlut.edu.cn)}
\thanks{Xinzhuo Yu and Jin Zhan are with the School of Computer Science and Technology, Dalian University of Technology, Dalian 116081, China. (e-mail: yxz\_3132@mail.dlut.edu.cn; zhanjin@dlut.edu.cn)}}
\markboth{IEEE Transactions on Multimedia,~Vol.~xx, No.~xx, Month~Year}%
{Author \MakeLowercase{\textit{et al.}}: Spectral Invariant Fusion and Refinement for Domain Generalized Semantic Segmentation}

\maketitle


\begin{abstract}
Domain generalized semantic segmentation is challenging because it requires accurate dense prediction on unseen domains without any access to target-domain data during training. Existing methods improve transferability through data diversification, invariant learning, or direct use of foundation models, but they often struggle to simultaneously suppress domain-sensitive interference and preserve fine-grained spatial cues. In this work, we propose SIFR, a frequency-aware dual-stream framework  that tackles domain shift at both the fusion and decoding stages. SIFR uses spectral modeling as a unified principle to regulate both heterogeneous prior interaction and dense decoder refinement. Instead of directly fusing frozen CLIP and DINOv2 features, SIFR performs spectral-guided invariant fusion to filter out domain-sensitive responses while preserving complementary semantic and structural information. In addition, SIFR incorporates a spectral decoder refinement module that stabilizes multi-scale decoder features through low-frequency statistics-based normalization and adaptive frequency selection, thereby reducing the propagation of residual domain bias in dense prediction. This design enables SIFR to jointly improve domain robustness and segmentation precision under severe distribution shifts. Extensive experiments on multiple DGSS benchmarks, including synthetic-to-real, real-to-real, and adverse-weather settings, demonstrate consistent improvements over previous state-of-the-art methods.


\end{abstract}



\section{Introduction}
Domain Generalized Semantic Segmentation (DGSS) aims to learn segmentation models from multiple source domains that generalize well to unseen target domains without any target-domain adaptation~\cite{choi2021robustnet,lee2022wildnet,yi2024learning,peng2022semantic,liao2026multi,wei2024stronger,zhang2026learning}. Compared with Domain Adaptive Semantic Segmentation~\cite{hoyer2022daformer,xue2025CACP,yu2025contrastive}, which addresses cross-domain discrepancies by leveraging unlabeled target-domain data during training, DGSS must confront domain shifts in a more challenging setting where no target-domain information is available at all. Consequently, a central challenge in DGSS is to achieve robustness to both low-level appearance changes and high-level semantic or contextual variations without compromising the accuracy and structural consistency of pixel-wise predictions~\cite{benigmim2024collaborating,choi2021robustnet,lee2022wildnet}.

\begin{figure*}[!t]
  \centering
  \includegraphics[width=\textwidth]{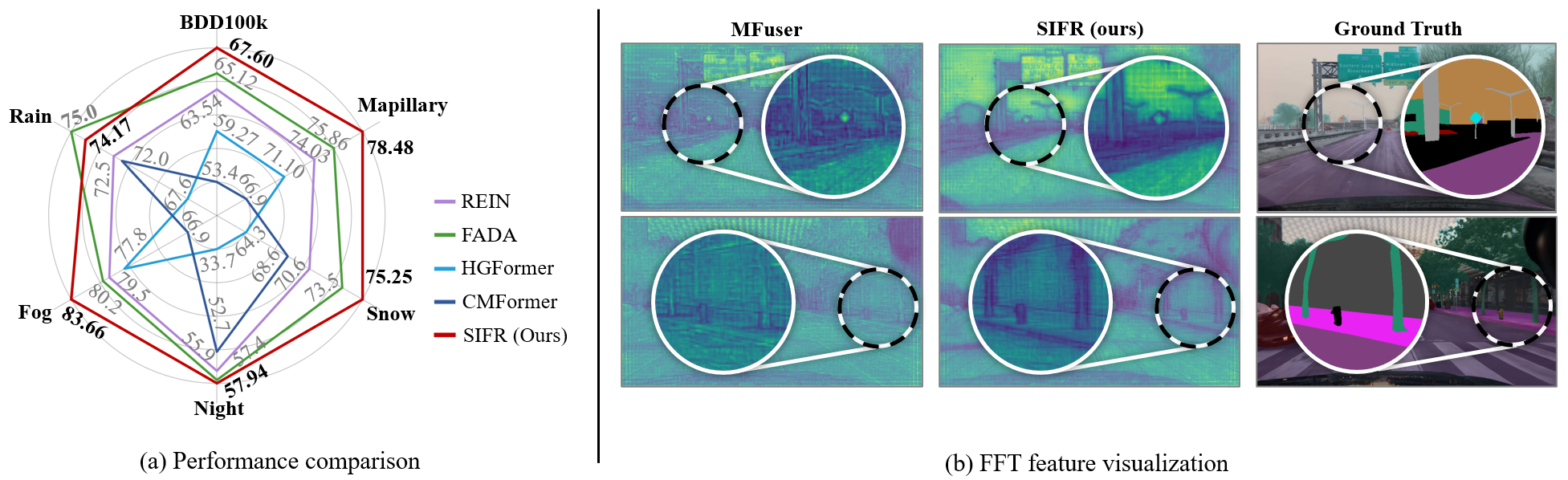}
    \put(-350,90){\scriptsize \cite{wei2024stronger}} 
    \put(-348,78){\scriptsize \cite{bi2024learning}} 
    \put(-335,67){\scriptsize \cite{ding2023hgformer}} 
    \put(-334,56){\scriptsize \cite{bi2024learning2}} 
    \put(-254,146){\scriptsize \cite{zhang2025mamba}} 
  \caption{\textbf{Representative characteristic of SIFR}. (a) Performance comparison with recent DGSS methods on representative real-to-real and adverse-weather benchmarks, where SIFR consistently achieves the best mIoU. (b) Feature visualization comparing MFuser~\cite{zhang2025mamba} and SIFR, where our method preserves clearer structures, suppresses domain-sensitive interference, and produces feature responses that better match the ground-truth semantics in challenging regions.}
  \label{fig:tissue}
\end{figure*}
Existing DGSS methods mainly improve generalization from two perspectives: data-level generalization and feature-level representation learning~\cite{rafi2024domain,huang2023style,lee2022wildnet,peng2022semantic}. The former increases source-domain diversity through perturbation, stylization, or synthesis, while the latter aims to learn domain-invariant representations across source domains~\cite{lee2022wildnet,yang2020fda,kim2023texture,peng2022semantic,yang2023generalized,huang2023style}. However, both directions face an inherent tension between domain invariance and spatial discriminability. Increasing appearance diversity or enforcing feature consistency can improve robustness to domain shifts, but may also weaken boundary, texture, and other fine-grained cues that are essential for accurate segmentation. Consequently, those methods still struggle to jointly achieve strong domain robustness and precise spatial prediction~\cite{rafi2024domain,peng2022semantic}. 

Recent advances in large-scale pretrained models have opened a promising direction for DGSS, as they provide more transferable and stable visual priors than conventionally trained backbones~\cite{wei2024stronger}. In particular, vision-language models (VLMs) capture rich high-level semantic knowledge, while vision foundation models (VFMs) offer stronger local structural cues and dense visual representations~\cite{radford2021learning,oquab2023dinov2,zhang2025mamba}. However, simply combining these heterogeneous priors is still insufficient for DGSS, as it faces two core challenges: \textbf{i) Cross-prior misalignment}. VLM and VFM are pretrained with different objectives, which leads to inconsistent feature geometries and domain responses and thus makes their interaction unstable under domain shifts. \textbf{ii) Domain appearance entanglement}. Despite their strong transferability, both priors can still encode domain-specific style, texture, and low-level statistics, so direct fusion may propagate domain-sensitive variations instead of isolating task-relevant cues.

To address these challenges, spectral modeling offers a complementary perspective. \textbf{i) Spectral alignment}. By organizing heterogeneous priors in the frequency domain, spectral modeling promotes more coherent interaction between semantic and structural cues across domains. \textbf{ii) Spectral disentanglement}. By separating domain-variant appearance components from task-relevant structural information, spectral modeling suppresses domain-sensitive responses while preserving the local details required for precise segmentation.

Inspired by these principles, we propose \textbf{SIFR}, a \textbf{S}pectral \textbf{I}nvariant \textbf{F}usion and \textbf{R}efinement framework for domain generalized semantic segmentation, which leverages heterogeneous large-scale priors while explicitly enhancing domain invariance. \textbf{i) Spectral-Guided Invariant Fusion (SGIF)}. SGIF employs frozen CLIP~\cite{radford2021learning} and DINOv2~\cite{oquab2023dinov2} as the VLM and VFM, where CLIP provides high-level semantic priors and DINOv2 captures fine-grained structural representations. By casting cross-prior interaction into the spectral domain, SGIF explicitly suppresses domain-variant appearance responses and enhances domain-stable semantic and structural cues. As qualitatively illustrated in Fig.~\ref{fig:tissue} (b), this spectral-guided invariant fusion produces clearer structural responses and better preserves semantically relevant regions under challenging domain shifts. Learnable query tokens and cross-attention~\cite{jaegle2021perceiver} are further introduced to promote effective semantic-visual fusion under reduced domain interference. \textbf{ii) Spectral Decoder Refinement (SDR)}. SDR performs decoder refinement via domain-stable normalization and frequency-guided enhancement module to curb the accumulation of domain-sensitive patterns, stabilize multi-scale features, and preserve informative components for precise segmentation. Overall, SIFR unifies spectral invariant fusion and refinement into a coherent framework that enhances generalization to unseen domains without sacrificing fine-grained spatial prediction.

We evaluate SIFR on multiple DGSS benchmarks spanning real-to-real and real-to-adverse-weather settings, the \textcolor{blue}{comparison results} of which is shown in Fig.~\ref{fig:tissue} (a). Furthermore, on the synthetic-to-real setting, SIFR achieves state-of-the-art performance, reaching 71.30 mIoU on Cityscapes~\cite{cordts2016cityscapes}, 62.54 on BDD100K~\cite{yu2020bdd100k}, and 71.96 on Mapillary~\cite{neuhold2017mapillary}, with gains of up to +1.32 points over previous methods. On the real-to-real setting, SIFR consistently outperforms strong baselines, obtaining 67.60 and 78.48 mIoU on BDD100K~\cite{yu2020bdd100k} and Mapillary~\cite{neuhold2017mapillary}, respectively. These consistent improvements suggest that SIFR effectively suppresses domain-sensitive interference while preserving transferable semantic and structural cues, thereby enhancing robustness under domain shift.

Our contributions can be summarized as follows:
\begin{itemize}
    \item We present SIFR, a frequency-aware dual-stream framework for domain generalized semantic segmentation that leverages frozen CLIP and DINOv2 as complementary semantic and structural priors, and introduces spectral modeling in both feature fusion and decoder refinement.
    \item We introduce a Spectral-Guided Invariant Fusion (SGIF) module that couples token-level cross-attention with frequency-domain refinement to promote complementary interaction between heterogeneous pretrained representations while suppressing domain-sensitive responses.
    \item We develop a Spectral Decoder Refinement (SDR) module, comprising domain-stable normalization and frequency-guided enhancement, to stabilize decoder features and preserve segmentation-relevant spectral components for robust dense prediction under domain shift.
    \item Extensive experiments on multiple DGSS benchmarks show that SIFR consistently improves cross-domain segmentation performance across synthetic-to-real, real-to-real, and adverse-weather settings.
\end{itemize}

\section{Related Work}
\subsection{Domain Generalized Semantic Segmentation}
Domain-generalized semantic segmentation (DGSS) studies how to learn segmentation models that generalize to unseen target domains without access to target-domain data during training. Existing research has explored this problem through multiple paradigms~\cite{benigmim2024collaborating,rafi2024domain,kim2022pin,choi2021robustnet,peng2022semantic}. Earlier efforts often improve domain robustness either by simulating domain variation during training, such as through meta-learning or source-domain diversification~\cite{kim2022pin,lee2022wildnet,yang2023generalized}, or by reducing domain-specific bias via feature alignment, normalization, whitening, and related representation learning techniques~\cite{choi2021robustnet,peng2022semantic,ahn2024style}. Beyond these data- and representation-oriented strategies, some studies further investigate architectural designs, including transformer-based models, to strengthen cross-domain feature modeling and dense prediction~\cite{ding2023hgformer,pak2024textual}. More recently, foundation models have introduced a new opportunity for DGSS, since large-scale pretraining can provide stronger and more transferable visual priors than conventionally trained backbones~\cite{wei2024stronger,benigmim2024collaborating,bi2024learning,zhang2025mamba}. Existing attempts have begun to exploit such priors through synthetic data generation~\cite{benigmim2024collaborating}, model adaptation~\cite{wei2024stronger,bi2024learning}, or semantic guidance~\cite{pak2024textual,hummer2023vltseg}. \textcolor{blue}{Beyond conventional static vision--language alignment,
SBGen~\cite{zheng2026learning} formulates the transformation from
domain-specific visual representations to domain-agnostic textual
semantics as a cross-modal Schr\"odinger Bridge, and models this
alignment as a stochastic semantic evolution process.} However, how to effectively utilize heterogeneous pretrained representations under domain shift, especially when different pretrained streams capture complementary yet unequally domain-sensitive cues, still remains to be explored.
\begin{figure*}[t]
    \centering
    \includegraphics[width=\textwidth]{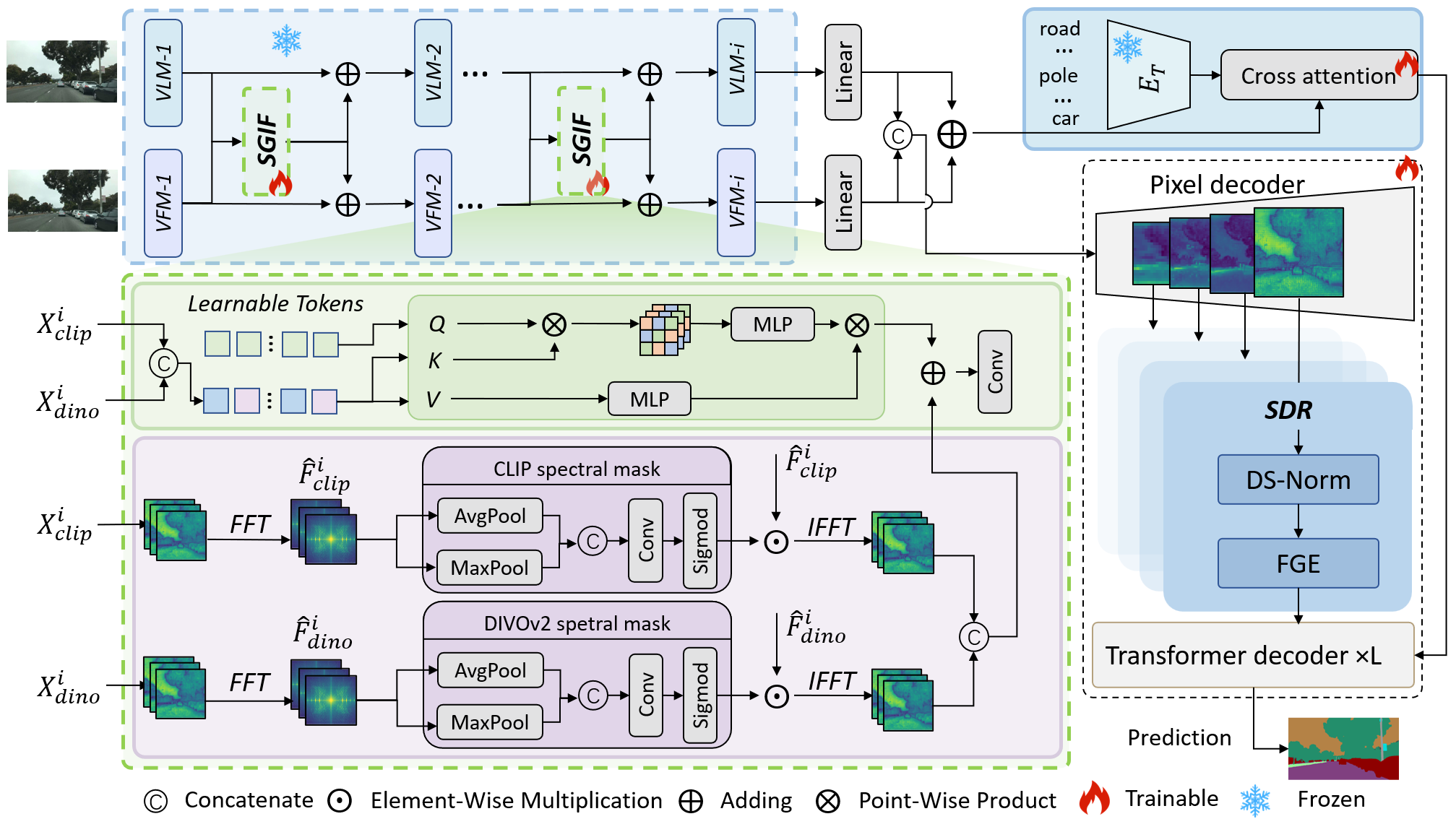}
    \caption{\textbf{Overview of the proposed framework}. Frozen VLM and VFM encoders extract hierarchical features, which are progressively fused by the proposed Spectral-Guided Invariant Fusion (SGIF) module via token-level cross-attention and frequency-domain refinement. The fused features are then passed to the pixel decoder and further refined by Spectral Decoder Refinement (SDR) module, including \textcolor{blue}{Domain-Stable Normalization (DS-Norm)} and Frequency-Guided Enhancement (FGE), before being decoded with text-guided semantic priors to generate robust segmentation masks under domain shift. }
    \label{fig:framework}
\end{figure*}

\subsection{Vision Foundation Models}
Vision foundation models (VFMs) have recently emerged as a powerful basis for visual transfer learning, primarily driven by large-scale pretraining on diverse image corpora using self-supervised objectives~\cite{caron2021emerging,fang2023eva,oquab2023dinov2}, masked image modeling~\cite{he2022masked}, and segmentation-oriented paradigms ~\cite{ravi2024sam}. Compared with conventionally supervised backbones, VFMs generally learn more generalizable and reusable representations, enabling strong transfer performance across a wide range of downstream tasks~\cite{he2022masked,fang2023eva,oquab2023dinov2}. Representative examples include DINO series~\cite{caron2021emerging,oquab2023dinov2}, iBOT, MAE~\cite{he2022masked} and EVA series~\cite{fang2023eva,fang2024eva,sun2023eva} which provide strong semantic and structural representations, as well as the SAM family~\cite{ravi2024sam,carion2025sam}, which brings foundation-scale modeling to segmentation.

Recent studies have also shown the relevance of VFMs to dense prediction. DINO~\cite{caron2021emerging} and iBOT~\cite{zhou2021ibot} suggest that self-supervised ViTs can encode explicit segmentation-related information and strong local semantic patterns, while 
ViT-Adapter~\cite{chen2022vision} demonstrates that pretrained ViTs can be effecti- vely adapted to semantic segmentation with lightweight task-specific designs. Nevertheless, existing efforts mostly emphasize architectural adaptation or downstream transfer, with limited attention paid to how domain-sensitive appearance statistics are entangled with transferable structural cues in pretrained dense representations. This leaves room for spectral modeling as a complementary mechanism for DGSS, since it can help expose and suppress domain-variant responses while preserving the structure-aware cues that VFMs already capture.

\subsection{Frequency Modeling Under Domain Shift}
Frequency-based modeling has become an increasingly important direction in cross-domain learning, since domain shift often affects not only spatial appearance but also frequency statistics~\cite{yang2020fda,huang2021fsdr,xu2021fourier,bi2024learning}. Existing studies mainly explore frequency information from two perspectives~\cite{yang2020fda,huang2021fsdr,xu2021fourier,bi2024learning}. One line of work applies frequency-domain perturbation, spectral substitution, or amplitude manipulation at the input level to expand source-domain diversity and alleviate style-related bias~\cite{yang2020fda,huang2021fsdr,xu2021fourier}. Another line of work performs frequency filtering or modulation at the feature level to improve the cross-domain robustness of learned representations~\cite{bi2024learning,chen2025frequency,chen2024frequency}. 
\textcolor{blue}{Most recently, HarDiff~\cite{bi2025simple} introduces the Discrete
Hartley Transform into a diffusion-based framework for generalizable
dense prediction. It exploits low-frequency structural priors from the
source domain during training and high-frequency detail guidance from
unseen-domain inputs during iterative sampling.}
{\color{blue}
Nevertheless, existing frequency-aware VFM methods mainly
focus on adapting a single pretrained visual stream. For
example, FADA~\cite{bi2024learning} decomposes VFM features
into low- and high-frequency branches through Haar wavelet
transformation, while SET~\cite{yi2024learning} processes the
amplitude and phase components of frozen VFM features with
learnable spectral tokens. In contrast, our method focuses on
frequency-aware interaction between heterogeneous VLM and VFM
representations rather than frequency adaptation within a
single VFM. Although MFuser~\cite{zhang2025mamba} also
integrates VLM and VFM features, it mainly employs Mamba-based
modules to model sequential and spatial interactions, without
explicitly regulating the domain-sensitive spectral responses
propagated from the two pretrained streams. Our SGIF instead
combines token-level cross-attention with stream-specific
adaptive spectral reweighting, and SIFR further extends
spectral regulation to multi-scale decoder refinement.
Therefore, SIFR differs from existing approaches in both the
representation scope and the stages at which spectral modeling
is performed.}


\section{Method}

\subsection{Overview}
As shown in Fig.~\ref{fig:framework}, our method adopts a unified frequency-aware segmentation framework that consists of hierarchical prior extraction, spectral-guided invariant fusion, and decoder-side refinement. Given an input image, frozen VLM and VFM encoders first extract multi-level features, where VLM provides high-level semantic priors and VFM offers fine-grained structural cues. To effectively exploit these heterogeneous representations under domain shift, we introduce a Spectral-Guided Invariant Fusion (SGIF) module during layer-wise feature interaction. Specifically, SGIF combines token-level cross-attention with a frequency-guided enhancement branch, which transforms features into the spectral domain, adaptively reweights different frequency components, and then projects them back to the spatial domain. In this way, the fusion process suppresses domain-sensitive appearance variations while preserving domain-stable semantic and structural information that is more beneficial for semantic segmentation.

The fused features are then fed into the pixel decoder and further refined by the proposed Spectral Decoder Refinement (SDR) module. Specifically, SDR module first applies low-frequency statistics-based normalization to mitigate residual domain-specific bias in decoder features, and then performs adaptive frequency selection to enhance segmentation-relevant spectral responses. The resulting multi-scale features are finally fed into the Transformer decoder, where they interact with text-guided semantic priors to progressively produce high-quality masks. \textcolor{blue}{At the framework level, SGIF, the pixel decoder, and SDR play
complementary rather than redundant roles. SGIF suppresses
domain-sensitive interference during heterogeneous feature
interaction and fusion. The pixel decoder subsequently
aggregates the SGIF-fused multi-level representations into
multi-scale dense features, providing spatially structured
visual representations for subsequent mask decoding. SDR then
mitigates the residual domain bias that remains or accumulates
during multi-scale feature aggregation and dense decoding.
Finally, the text-guided semantic priors complement the
SDR-refined visual representations by providing category-level
semantic guidance within the Transformer decoder, thereby
establishing more reliable pixel-to-class correspondences for
mask prediction.}

\subsection{Spectral-Guided Invariant Fusion}
In domain generalized semantic segmentation, pretrained VLMs~\cite{radford2021learning} and VFMs~\cite{oquab2023dinov2} often exhibit domain preferential responses learning from their pretraining data. Therefore, directly combining their features can introduce domain-sensitive patterns during feature interaction and fusion. To address this issue, we formulate feature fusion as a dual-branch process, as shown in Fig.~\ref{fig:framework}, consisting of a layer-wise interaction branch and a frequency-guided enhancement branch. The former captures complementary semantic interactions between the two pretrained representations, while the latter injects spectral priors to suppress domain-sensitive interference and preserve structurally stable cues.


Given the features extracted from the $i$-th layer of VLM and VFM, we denote them as
$X_{\text{vlm}}^{i} \in \mathbb{R}^{B \times L_c \times D}$ and
$X_{\text{vfm}}^{i} \in \mathbb{R}^{B \times L_d \times D}$,
where $B$ is the batch size, $L_c$ and $L_d$ denote the token numbers, and $D$ is the shared embedding dimension. 
To capture complementary interactions between the two branches, we first concatenate their features along the sequence dimension:
\begin{equation}
X_{\text{cat}}^{i} = \text{Concat}\left(X_{\text{vlm}}^{i}, X_{\text{vfm}}^{i}\right).
\end{equation}

We then introduce a learnable query token $T^{i} \in \mathbb{R}^{B \times L_q \times D}$ and perform cross-attention over the concatenated features:
\begin{equation}
X_{\text{attn}}^{i} = \text{CrossAttn}\left(T^{i}, X_{\text{cat}}^{i}, X_{\text{cat}}^{i}\right),
\end{equation}
where $X_{\text{attn}}^{i}$ denotes the $i$-th layer fusion result and \textcolor{blue}{$L_q$ denotes the number of learnable query tokens.}

In parallel, we construct a frequency-guided enhancement branch. Specifically, features are reshaped into 2D feature maps,
$F_{\text{vlm}}^{i} \in \mathbb{R}^{B \times D \times H_c \times W_c}$ and
$F_{\text{vfm}}^{i} \in \mathbb{R}^{B \times D \times H_d \times W_d}$. 
We then transform them into the frequency domain using Fourier transform operation $\mathcal{F}(\cdot)$, yielding
$\hat{F}_{\text{vlm}}^{i}=\mathcal{F}(F_{\text{vlm}}^{i})$ and
$\hat{F}_{\text{vfm}}^{i}=\mathcal{F}(F_{\text{vfm}}^{i})$.
Based on the spectral representations, we generate the adaptive spectral masks $M_{\text{vlm}}^{i}$ and $M_{\text{vfm}}^{i}$ through a lightweight refinement module. \textcolor{blue}{For each branch, the spectral feature is first summarized by channel-wise average and maximum pooled descriptors computed from the magnitude spectrum at each frequency location. These descriptors preserve the spatial frequency resolution and are concatenated and passed through a convolutional layer followed by a sigmoid function to produce the corresponding spectral mask.}
The spectral masks are applied to reweight the frequency responses:
\begin{equation}
\tilde{F}_{\text{vlm}}^{i} = \hat{F}_{\text{vlm}}^{i} \odot M_{\text{vlm}}^{i}, \qquad
\tilde{F}_{\text{vfm}}^{i} = \hat{F}_{\text{vfm}}^{i} \odot M_{\text{vfm}}^{i},
\end{equation}
where $\odot$ denotes element-wise multiplication. After inverse Fourier transform, the enhanced feature maps are transformed back into token representations, denoted as $X_{\text{freq}}^{i}$.

Finally, we combine the token-level fusion result and the frequency-guided enhancement through residual addition, followed by a convolutional layer to obtain the final fused representation:
\begin{equation}
X_{\text{fuse}}^{i} = \text{Conv}\left(X_{\text{attn}}^{i} + X_{\text{freq}}^{i}\right).
\end{equation}

In this way, the proposed fusion strategy integrates the complementary semantic interactions captured by the cross-attention branch with the domain-robust structural cues provided by the frequency-guided branch. As a result, it yields more discriminative and robust intermediate features for the subsequent decoder under domain shift.


\begin{figure}[t]
    \centering
    \includegraphics[width=0.5\textwidth]{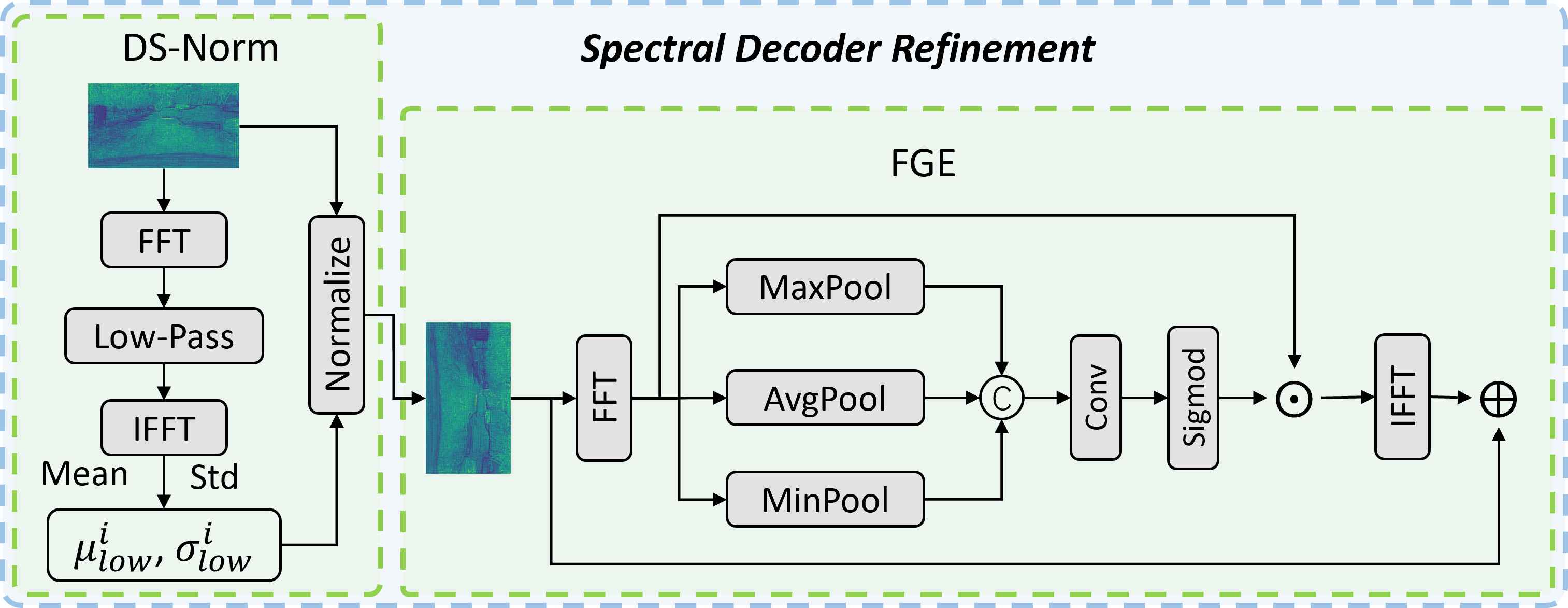}
    \caption{\textbf{Paradigm of the Spectral Decoder Refinement (SDR) module}. For each decoder feature, DS-Norm first normalizes the feature using low-frequency statistics obtained via FFT, low-pass filtering, and IFFT, while FGE then adaptively enhances segmentation-relevant frequency responses through pooled spectral descriptors and inverse reconstruction.}
    \label{fig:model2}
\end{figure}
\subsection{Spectral Decoder Refinement}

{\color{blue}
Although the encoder-stage SGIF reduces domain-sensitive interference during heterogeneous prior fusion, residual appearance deviations may still remain in the fused representations and propagate through subsequent multi-scale decoding. The key motivation of SDR is to decouple residual domain-bias suppression from segmentation-relevant spectral preservation during dense prediction. Since global and smoothly varying domain changes, such as illumination, contrast, weather, and style, are more strongly reflected in low-frequency feature statistics, DS-Norm estimates normalization statistics from the low-frequency component while applying normalization to the original full-spectrum feature. This suppresses global statistical deviations without explicitly discarding fine-grained spatial responses. However, normalization alone cannot determine which frequency components are informative for different images and feature scales. Therefore, FGE further performs data-dependent spectral reweighting to adaptively emphasize segmentation-relevant responses. In this way, SDR follows a suppress-then-select principle: DS-Norm stabilizes decoder feature statistics, whereas FGE adaptively preserves and enhances informative spectral components for robust mask prediction.
As shown in Fig.~\ref{fig:framework}, SDR is inserted between the pixel decoder and the Transformer decoder. Fig.~\ref{fig:model2} details the internal suppress-then-select design of SDR.}

Let $\{D^i\}_{i=1}^{4}$ denote the four multi-scale output features of the pixel decoder, where $D^i \in \mathbb{R}^{B\times C\times H_i\times W_i}$. For each scale, we first extract its low-frequency component in the frequency domain. 
Specifically, we transform $D^i$ into the spectral domain, apply a low-pass filter, and project the filtered response back to the spatial domain, yielding the low-frequency feature $D_{\mathrm{low}}^i$. \textcolor{blue}{The size of the retained low-frequency region is controlled by a
filter ratio $\rho$.} Based on $D_{\mathrm{low}}^i$, we compute the channel-wise mean and standard deviation:
\begin{equation}
\mu_{\mathrm{low}}^i=\mathrm{Mean}(D_{\mathrm{low}}^i), \qquad
\sigma_{\mathrm{low}}^i=\mathrm{Std}(D_{\mathrm{low}}^i).
\end{equation}

These low-frequency statistics are then used to normalize the original decoder feature:
\begin{equation}
\tilde{D}^i=\frac{D^i-\mu_{\mathrm{low}}^i}{\sigma_{\mathrm{low}}^i+\epsilon},
\end{equation}
where $\epsilon$ is a small constant for numerical stability.

Built upon the normalized feature $\tilde{D}^i$, we further perform frequency-guided enhancement. 
We first transform $\tilde{D}^i$ into the frequency domain and denote its spectral representation as:
\begin{equation}
\hat{\tilde{D}}^i = \mathcal{F}(\tilde{D}^i).
\end{equation}


\textcolor{blue}{Then, we generate the frequency-guided mask $M^i$ from $\hat{\tilde{D}}^i$ using a lightweight refinement module. Specifically, for each frequency location, we compute channel-wise pooled spectral descriptors from the magnitude spectrum of $\hat{\tilde{D}}^i$, including average, maximum, and minimum responses across the channel dimension. Different from spatial pooling with a fixed local window, these operations summarize cross-channel spectral statistics while preserving the spatial frequency resolution. The resulting descriptors are concatenated and fed into a lightweight convolutional layer followed by a sigmoid function to produce $M^i$.} The mask is then applied to reweight the normalized spectral representation:
\begin{equation}
\hat{D}_{\mathrm{enh}}^i = M^i \odot \hat{\tilde{D}}^i.
\end{equation}

Finally, the enhanced decoder feature is obtained by inverse Fourier transform:
\begin{equation}
D_{\mathrm{enh}}^i=\mathcal{F}^{-1}(\hat{D}_{\mathrm{enh}}^i).
\end{equation}

In this way, the proposed decoder refinement suppresses residual domain-sensitive statistical deviations while further emphasizing structurally informative spectral responses. As a result, it provides more robust multi-scale features for the subsequent transformer decoder and improves mask prediction under domain shift.

\subsection{Training Objective}
We optimize the proposed framework using the standard segmentation objective together with a pixel-level vision-language alignment objective. 
For segmentation, following Mask2Former~\cite{cheng2022masked}, the loss is defined as:
\begin{equation}
\mathcal{L}_{\mathrm{seg}}
=
\lambda_{\mathrm{bce}}\mathcal{L}_{\mathrm{bce}}
+
\lambda_{\mathrm{dice}}\mathcal{L}_{\mathrm{dice}}
+
\lambda_{\mathrm{cls}}\mathcal{L}_{\mathrm{cls}},
\end{equation}
where \(\mathcal{L}_{\mathrm{bce}}\) and \(\mathcal{L}_{\mathrm{dice}}\) denote the binary cross-entropy loss and Dice loss for mask prediction, respectively, and \(\mathcal{L}_{\mathrm{cls}}\) is the classification loss for query prediction.

{\color{blue}

In addition, we introduce a pixel-text alignment loss
$\mathcal{L}_{\mathrm{align}}$ to enforce semantic consistency
between pixel-level visual representations and textual class
embeddings. For each semantic class $k$, we construct its text
prompt by placing $M$ learnable context tokens
$\{p_m\}_{m=1}^{M}$ between the start token $\mathrm{SOS}$
and the tokenized class name $c_k$, followed by the end token
$\mathrm{EOS}$. The CLIP text encoder is kept frozen, while
the context tokens are optimized during training.

Given the normalized pixel-level visual feature
$\hat{\mathbf{v}}_i$ at pixel $i$ and the normalized text
embedding $\hat{\mathbf{t}}_k$ of class $k$, we compute their
similarity score as the inner product
$\hat{\mathbf{v}}_i^{\top}\hat{\mathbf{t}}_k$ divided by the
temperature $\tau$, which is fixed to $0.07$. The resulting
dense similarity scores are supervised by the ground-truth
semantic labels using pixel-wise cross-entropy:
\begin{equation}
\mathcal{L}_{\mathrm{align}}
=
-\frac{1}{|\Omega|}
\sum_{i \in \Omega}
\log
\frac{\exp(s_{i,y_i})}
{\sum_{k=1}^{K}\exp(s_{i,k})},
\end{equation}
where $K$ denotes the number of semantic classes, $y_i$ is
the ground-truth label of pixel $i$, and $\Omega$ is the set
of valid pixels. Pixels with the ignore label are excluded.

The overall training objective is defined as
\begin{equation}
\label{eq:total_loss}
\mathcal{L}_{\mathrm{total}}
=
\mathcal{L}_{\mathrm{seg}}
+
\lambda_{\mathrm{align}}
\mathcal{L}_{\mathrm{align}},
\end{equation}
where $\lambda_{\mathrm{align}}$ is set to $1.0$ in all
experiments.
}

\begin{figure*}[t]
    \centering
    \includegraphics[width=1.01\textwidth]{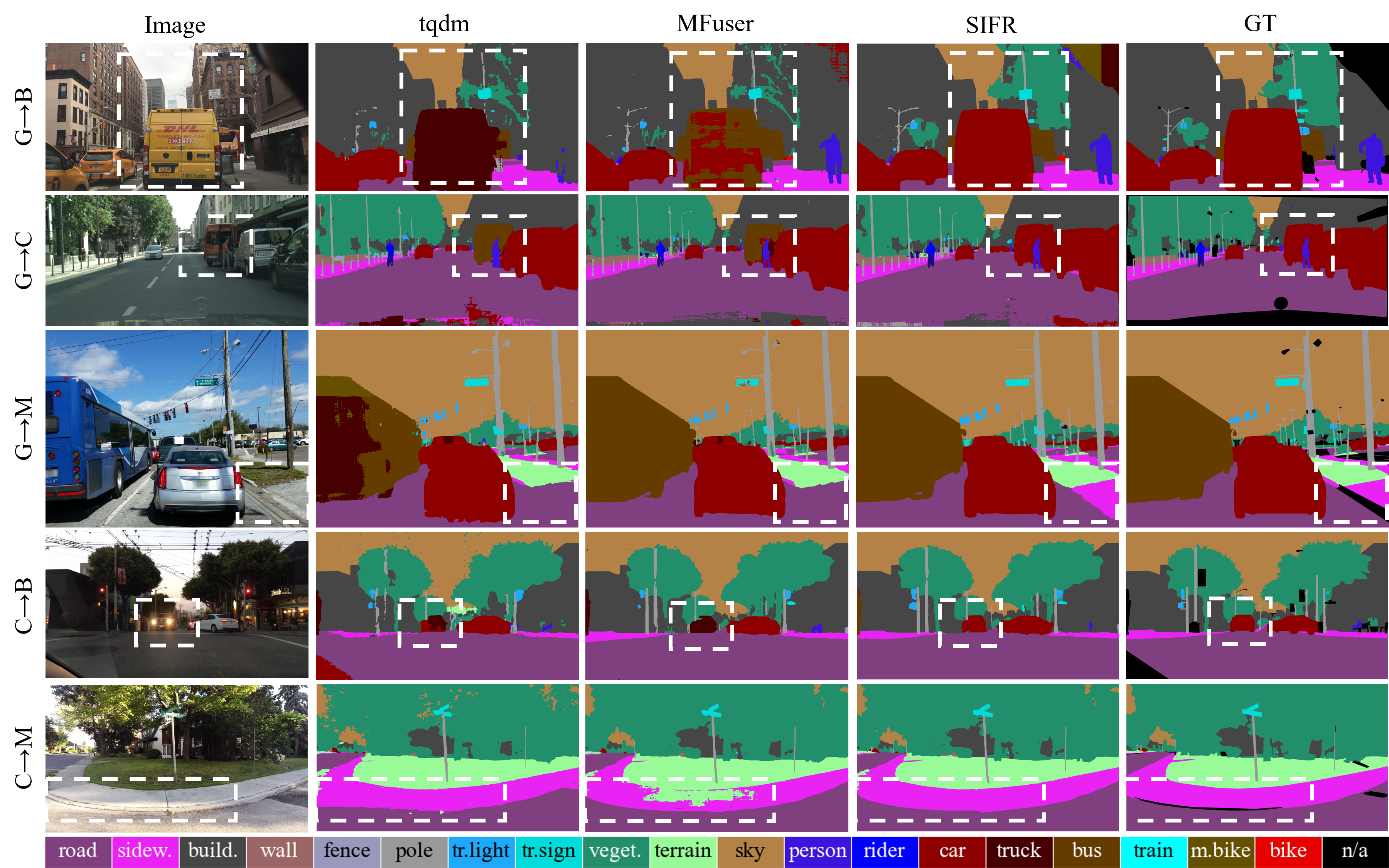}
    \vspace{-4mm}
    \caption{\textbf{Qualitative comparison of segmentation results across five source-to-target settings: $ G\to{B} $, $ G\to{C} $, $ G\to{M} $, $ C\to{B} $, and $ C\to{M} $.} From left to right are the input image, the predictions of tqdm~\cite{pak2024textual}, MFuser~\cite{zhang2025mamba}, our SIFR, and the ground truth . The white boxes highlight challenging regions with small objects, thin structures, and ambiguous boundaries. Compared with previous methods, SIFR yields more complete predictions and better preserves fine structures, producing results closer to the ground truth.}
    \label{fig:vision}
    \vspace{-2mm}
\end{figure*}

\section{Experiments}

\subsection{Experiment Settings}
\textbf{Datasets and Metrics}. We evaluate our method under synthetic-to-real, real-to-real, and real-to-adverse-weather DGSS settings on five driving-scene benchmarks that share 19 semantic categories. \textcolor{blue}{GTA5}~\cite{richter2016playing} is a large-scale synthetic dataset containing 24,966 densely annotated images with a fixed resolution of 1914×1052, and is widely used as a source domain for cross-domain semantic segmentation. Cityscapes~\cite{cordts2016cityscapes} is a standard real-world benchmark for urban scene understanding, containing 2,975 training images, 500 validation images, and 1,525 testing images, with a fixed resolution of 2048×1024. BDD100K~\cite{yu2020bdd100k} is a diverse real-world driving dataset with images of fixed resolution 1280×720; for semantic segmentation, it provides 10K annotated images and is commonly organized into 7,000 training, 1,000 validation, and 2,000 test images. Mapillary~\cite{neuhold2017mapillary} is a large-scale street-scene dataset containing 18,000 training images, 2,000 validation images, and 5,000 test images; unlike Cityscapes~\cite{cordts2016cityscapes} and BDD100K~\cite{yu2020bdd100k}, its images are high-resolution but come in diverse sizes and capture conditions. ACDC~\cite{sakaridis2021acdc} is designed for adverse-condition semantic segmentation and contains 1,600 training images, 406 validation images, and 2,000 test images under four weather conditions, \textit{i.e.} fog, rain, snow, and nighttime, making it a challenging benchmark for evaluations under severe domain shifts. Following prior DGSS works~\cite{wei2024stronger,zhang2025mamba}, we adopt mean Intersection over Union (mIoU) as the evaluation metric. 

\textbf{Implementation Details}. \textcolor{blue}{We adopt
CLIP~\cite{radford2021learning} as the vision-language model,
DINOv2~\cite{oquab2023dinov2} as the visual foundation model,
and Mask2Former~\cite{cheng2022masked} as the segmentation
head. During training, the CLIP visual encoder, CLIP text
encoder, and DINOv2 backbone are kept frozen, and their
parameters are not updated. Optimization is performed over the
parameters of the SGIF modules and their learnable query tokens,
the feature projection and fusion layers, the pixel decoder, the
learnable layers within SDR, the text-guided cross-attention
module, the Transformer decoder, and the classification and
mask prediction heads. The learnable context tokens used to
construct the class prompts are also optimized, while the CLIP
text encoder remains frozen. All trainable parameters are jointly
optimized in a single end-to-end training stage using the total
objective in Eq.(\ref{eq:total_loss}). No stage-wise pretraining or alternating
optimization is adopted.}
All input images are resized to 512 × 512. We train the model with a batch size of 2 on each GPU. We optimize the model using AdamW with a base learning rate of 1e-4 and a weight decay of 1e-4. The learning rate follows a poly decay schedule with a power of 0.9 and a minimum learning rate of 1e-6, together with a linear warm-up over the first 1900 iterations using a warm-up ratio of 1e-6. The model is trained for 30,000 iterations. During inference, we adopt sliding-window evaluation with a crop size of 512 × 512 and a stride of 341 × 341. \textcolor{blue}{Unless otherwise specified, we use $L_q=64$ learnable query
tokens in SGIF and set the low-pass filter ratio in DS-Norm to
$\rho=0.25$.} All experiments are implemented based on MMSegmentation and conducted on two RTX 4090 GPUs.

\subsection{\textcolor{blue}{Comparison with State-of-the-Art Methods}}
We compare SIFR with representative domain generalized semantic segmentation (DGSS) methods under three evaluation settings: synthetic-to-real generalization from \textcolor{blue}{GTA5}~\cite{richter2016playing} to Cityscapes~\cite{cordts2016cityscapes}, BDD100K~\cite{yu2020bdd100k}, and Mapillary~\cite{neuhold2017mapillary} ($G \to {C,B,M}$), real-to-real generalization from Cityscapes~\cite{cordts2016cityscapes} to BDD100K~\cite{yu2020bdd100k} and Mapillary~\cite{neuhold2017mapillary} ($C \to {B,M}$), and real-to-adverse-weather generalization from Cityscapes~\cite{cordts2016cityscapes} to ACDC~\cite{sakaridis2021acdc} ($C \to A$). Since our method is built upon heterogeneous pretrained priors, we place particular emphasis on recent foundation-model-based approaches, including CLOUDS~\cite{benigmim2024collaborating}, VLTSeg~\cite{hummer2023vltseg}, REIN~\cite{wei2024stronger}, SET~\cite{yi2024learning}, MFuser~\cite{zhang2025mamba}, DepthForge~\cite{chen2025stronger}, and tqdm~\cite{pak2024textual}. For completeness, we also include several conventional DGSS methods to provide a broader comparison across different methodological paradigms.


\textbf{Real-to-Real Generalization}.
Tab.~\ref{tab:C-BM} reports the comparisons under the real-to-real setting. SIFR achieves the best performance on both $C \to B$ and $C \to M$, demonstrating that the proposed spectral modeling strategy improves transferability across real-world domains. In particular, compared with the strong baseline MFuser~\cite{zhang2025mamba}, SIFR improves mIoU by +2.02 on $C \to B$ and +0.38 on $C \to M$, yielding an average gain of +1.20. Moreover, although DepthForge~\cite{chen2025stronger} additionally leverages depth information, SIFR still surpasses it by +1.41 on $C \to B$ and +2.55 on $C \to M$, corresponding to an average improvement of +1.98. These results indicate that SIFR more effectively exploits complementary semantic and structural priors for robust real-to-real generalization.

\textbf{Synthetic-to-Real Generalization}.
The results in Tab.~\ref{tab:G2CBM} further verify the effectiveness of SIFR under the more challenging synthetic-to-real setting. SIFR obtains the best performance on all three target domains, namely $G \to C$, $G \to B$, and $G \to M$. Relative to MFuser~\cite{zhang2025mamba}, SIFR improves performance by +0.06, +1.46, and +0.82 mIoU, respectively. The largest improvement on BDD100K~\cite{yu2020bdd100k} suggests that SIFR is particularly effective when the target domain exhibits larger appearance and context variations.


\textbf{Real-to-Adverse-Weather Generalization}. 
As reported in Tab.~\ref{tab: C2A}, SIFR achieves the best overall performance on Cityscapes~\cite{cordts2016cityscapes} $\to$ ACDC~\cite{sakaridis2021acdc}, showing stronger robustness under adverse weather. It reaches 75.3 mIoU on Snow, 57.9 on Night, and 83.7 on Fog, outperforming previous methods on all three conditions. Compared with FADA~\cite{bi2024learning}, SIFR improves by +1.8, +0.5, and +3.5 mIoU on Snow, Night, and Fog, while remaining competitive on Rain. This suggests that SIFR better preserves domain-invariant cues under severe weather-induced shifts.

\textbf{Qualitative Results Analysis.} 
Fig.~\ref{fig:vision} further illustrates the qualitative superiority of SIFR on five cross-domain settings spanning synthetic-to-real and real-to-real generalization. Compared with tqdm~\cite{pak2024textual} and MFuser~\cite{zhang2025mamba}, SIFR produces masks that are visually closer to the ground truth, especially in the white-boxed regions containing small objects, thin structures, and ambiguous boundaries. On $G \to B$ and $G \to  C$, our method better preserves object completeness for vehicles and roadside structures while reducing fragmented predictions. On $G \to M$ and $C \to M$, SIFR yields cleaner layouts for roads, sidewalks, and vegetation, with fewer category confusions. On the challenging $C \to B$ case, it also recovers clearer scene boundaries under low-light conditions. These observations confirm that spectral fusion and decoder refinement improve both structural consistency and semantic precision.

\begin{table}[!tbp]
    \centering
    \scriptsize
    \setlength{\abovecaptionskip}{0.1cm}
    \setlength{\belowcaptionskip}{0.1cm}
    \setlength{\tabcolsep}{4pt}
    \renewcommand{\arraystretch}{1.1}
    \vspace{-5mm}
    \caption{Quantitative comparisons under the \textit{Cityscapes $\rightarrow$ BDD100K + Mapillary} generalization setting.}
    \label{tab:C-BM}
    
    \resizebox{0.49\textwidth}{!}{
    \begin{tabular}{l|c|cc|c}  
        \toprule
        \textbf{Method} & \textbf{Proc. \& Year} & \textbf{BDD100k} & \textbf{Mapillary} & \textbf{Avg} \\
        \midrule
        \multicolumn{2}{l}{\textit{ResNet based:}} \\
        IBN~\cite{pan2018two} & ECCV2018 & 48.56 & 57.04 & 52.80 \\
        IW~\cite{pan2019switchable} & CVPR2019 & 48.49 & 55.82 & 52.16 \\
        Itenorm~\cite{huang2019iterative} & CVPR2019 & 49.23 & 56.26 & 52.75 \\
        DRPC~\cite{yue2019domain} & ICCV2019 & 49.86 & 56.34 & 53.10 \\
        GTR~\cite{peng2021global} & TIP2021 & 50.75 & 57.16 & 53.96 \\
        RobustNet~\cite{choi2021robustnet} & CVPR2021 & 50.73 & 58.64 & 54.69 \\
        WildNet~\cite{lee2022wildnet} & CVPR2022 & 50.94 & 58.79 & 54.87 \\
        SAW~\cite{peng2022semantic} & CVPR2022 & 52.95 & 59.81 & 56.38 \\
        BlindNet~\cite{ahn2024style} & CVPR2024 & 51.84 & 60.18 & 56.01 \\
        \midrule
        \multicolumn{2}{l}{\textit{Transformer based:}} \\
        HGFormer~\cite{ding2023hgformer} & CVPR2023 & 53.40 & 66.90 & 60.15 \\
        CMFormer~\cite{bi2024learning2} & AAAI2024 & 59.27 & 71.10 & 65.19 \\
        \midrule
        \multicolumn{2}{l}{\textit{VFM based:}} \\
        REIN~\cite{wei2024stronger} & CVPR2024 & 63.54 & 74.03 & 68.79 \\
        FADA~\cite{bi2024learning}  & NeurIPS2024 & 65.12 & 75.86 & 70.49 \\
        SET~\cite{yi2024learning} & ACM MM & 65.07 &75.76 & 70.37\\
        VLTSeg~\cite{hummer2023vltseg}& ACCV2024 & 64.40 &76.40 & 70.40\\
        tqdm~\cite{pak2024textual} & ECCV2024 & 64.72 & 76.15 & 70.44\\
        DPMFormer~\cite{jeon2025exploiting} & ICCV2025 & 64.20 & 76.67 & 70.44\\
        DepthForge~\cite{chen2025stronger} & ICCV2025 & 66.19 & 75.93 & 71.06 \\  
        Causal-Tune~\cite{zhang2026causal} & AAAI2026 & 66.28 & 76.05 & 71.16 \\ 
        MFuser~\cite{zhang2025mamba} & CVPR2025 & 65.58 & 78.10 & 71.84 \\ 
        SBGen~\cite{zheng2026learning}&NeurIPS2026  & 66.03 & 77.90 & 71.97 \\ 
        \rowcolor{gray!20}
       \textbf{SIFR(ours)} & - & \textbf{67.60} & \textbf{78.48} & \textbf{73.04}\\
        \bottomrule
    \end{tabular}
    }
\end{table}

{
\setlength{\abovecaptionskip}{0.1cm}
\setlength{\belowcaptionskip}{0.1cm}
\begin{table}[!t]
    \centering
    \caption{Quantitative Comparisons under the \textit{\textcolor{blue}{GTA5} $\rightarrow$ Cityscapes + BDD100K + Mapillary} generalization setting.}
    \setlength{\tabcolsep}{4pt}
    \renewcommand{\arraystretch}{1.1}
    \large
    \scalebox{0.65}{
    \begin{tabular}{l|c|cccc} 
        \toprule
        \textbf{Method} & \textbf{Proc. \& Year} & \textbf{Cityscapes} & \textbf{BDD100k} & \textbf{Mapillary} & \textbf{Avg} \\
        \midrule
        \multicolumn{2}{l}{\textit{ResNet based:}} \\
        Iternorm~\cite{huang2019iterative} & CVPR2019 & 31.81 & 32.70 & 33.88 & 32.80\\
        DRPC~\cite{yue2019domain} & ICCV2019 & 37.42 & 32.14 & 34.12 & 34.56 \\
        GTR~\cite{peng2021global}  & TIP2021 & 37.53 & 33.75 & 34.52 & 35.27 \\
        RobustNet~\cite{choi2021robustnet} & CVPR2021 & 36.58 & 35.20 & 40.33 & 37.37 \\
        AdvStyle~\cite{zhong2022adversarial}  & NeurIPS2022 & 39.62 & 35.54 & 37.00 & 37.39 \\
        SAW~\cite{peng2022semantic}  & CVPR2022 & 39.75 & 37.34 & 41.86 & 39.65 \\
        DIRL~\cite{xu2022dirl}  & AAAI2022 & 41.04 & 39.15 & 41.60 & 40.60 \\
        SHADE~\cite{zhao2022style} & ECCV2022 & 44.65 & 39.28 & 43.34 & 42.42 \\
        WildNet~\cite{lee2022wildnet}  & CVPR2023 & 44.62 & 38.42 & 46.09 & 43.04 \\
        SPC~\cite{huang2023style} & CVPR2023 & 44.10 & 40.46 & 45.51 & 43.36 \\
        BlindNet~\cite{ahn2024style}  & CVPR2024 & 45.72 & 41.32 & 47.08 & 44.71 \\
        TLDR~\cite{kim2023texture} &ICCV2023 &47.58 & 44.88 & 48.80 &47.08 \\
        FAMix~\cite{fahes2024simple} & CVPR2024 & 49.47& 46.40 &51.97 &49.28\\
        \midrule
        \multicolumn{2}{l}{\textit{Transformer based:}} \\
        CMFormer~\cite{bi2024learning2} & AAAI2024 & 55.31 & 49.91 & 60.09 & 55.10 \\
        \midrule
        \multicolumn{2}{l}{\textit{VFM based:}} \\
        DIDEX~\cite{niemeijer2024generalization} & WACV2024 & 62.00 & 54.30 & 63.00 & 59.77 \\
        CLOUDS~\cite{benigmim2024collaborating} & - & 60.20 & 57.40 &67.00&61.53  \\
        VLTSeg~\cite{hummer2023vltseg}& ACCV2024 & 65.60 &58.40 & 66.50 &63.50 \\
        REIN~\cite{wei2024stronger} & CVPR2024 & 66.40 & 60.40 & 66.10 & 64.30 \\
        SET~\cite{yi2024learning} & ACMMM2024 & 68.06 &61.64 & 67.68&65.79\\
        tqdm~\cite{pak2024textual} & ECCV2024 & 68.88 & 59.18 & 70.10 & 66.05\\
        FADA~\cite{bi2024learning}  & NeurIPS2024 & 68.23 & 61.94 & 68.09 & 66.09 \\
        HarDiff~\cite{bi2025simple} & ICCV2025 & 62.00 & 58.90 & 68.80 & 63.20 \\
        MFuser~\cite{zhang2025mamba}  & CVPR2025 & 71.24 & 61.08 & 71.14 & 67.82 \\
        SBGen~\cite{zheng2026learning} & NeurIPS2026  & 71.24 & 62.26 & 71.91 & 68.47 \\
        \rowcolor{gray!20}
        \textbf{SIFR} & - & \textbf{71.30} &  \textbf{62.54} &\textbf{71.96}  & \textbf{68.60} \\
        \bottomrule
    \end{tabular}}
    \label{tab:G2CBM}
\end{table}
}

\begin{table}[!t]
    \centering
    \setlength{\abovecaptionskip}{0.1cm}
    \setlength{\belowcaptionskip}{0.2cm}
    \setlength{\tabcolsep}{2pt}
    \caption{Quantitative comparisons under the \textit{Cityscapes $\rightarrow$ ACDC} adverse-weather generalization setting.}
    \tiny
    \renewcommand{\arraystretch}{1.1}
    \resizebox{0.45\textwidth}{!}{
    \begin{tabular}{l|c|cccc}
        \toprule
        \textbf{Method} & \textbf{Proc. \& Year} & \textbf{Snow} & \textbf{Night} & \textbf{Fog} & \textbf{Rain} \\
        \midrule
        \multicolumn{2}{l}{\textit{ResNet based:}} \\
        IBN~\cite{pan2018two} & ECCV2018 & 49.6 & 21.2 & 63.8 & 50.4 \\
        Itenorm~\cite{huang2019iterative} & CVPR2019 & 49.9 & 23.8 & 63.3 & 50.1 \\
        IW~\cite{pan2019switchable} & CVPR2019 & 47.6 & 21.8 & 62.4 & 52.4 \\
        RobustNet~\cite{choi2021robustnet} & CVPR2021 & 49.8 & 24.3 & 64.3 & 56.0 \\
        WildNet~\cite{lee2022wildnet} & CVPR2022 & 28.4 & 12.7 & 41.2 & 34.2 \\
        \midrule
        \multicolumn{2}{l}{\textit{Transformer based:}} \\
        HGFormer~\cite{ding2023hgformer} & CVPR2023 & 68.6 & 52.7 & 69.9 & 72.0 \\
        CMFormer~\cite{bi2024learning2} & AAAI2024 & 64.3 & 33.7 & 77.8 & 67.6 \\
        \midrule
        \multicolumn{2}{l}{\textit{VFM based:}} \\
        REIN~\cite{wei2024stronger} & CVPR2024 & 70.6 & 55.9 & 79.5 & 72.5 \\
        FADA~\cite{bi2024learning}  & NeurIPS2024 & 73.5 & 57.4 & 80.2 & 75.0 \\
        \rowcolor{gray!20}
        \textbf{SIFR}&-& 75.3&57.94&83.7&74.2\\
        \bottomrule
    \end{tabular}
    }
    \label{tab: C2A}
\end{table}


\begin{figure*}[t]
    \centering
    \includegraphics[width=\textwidth]{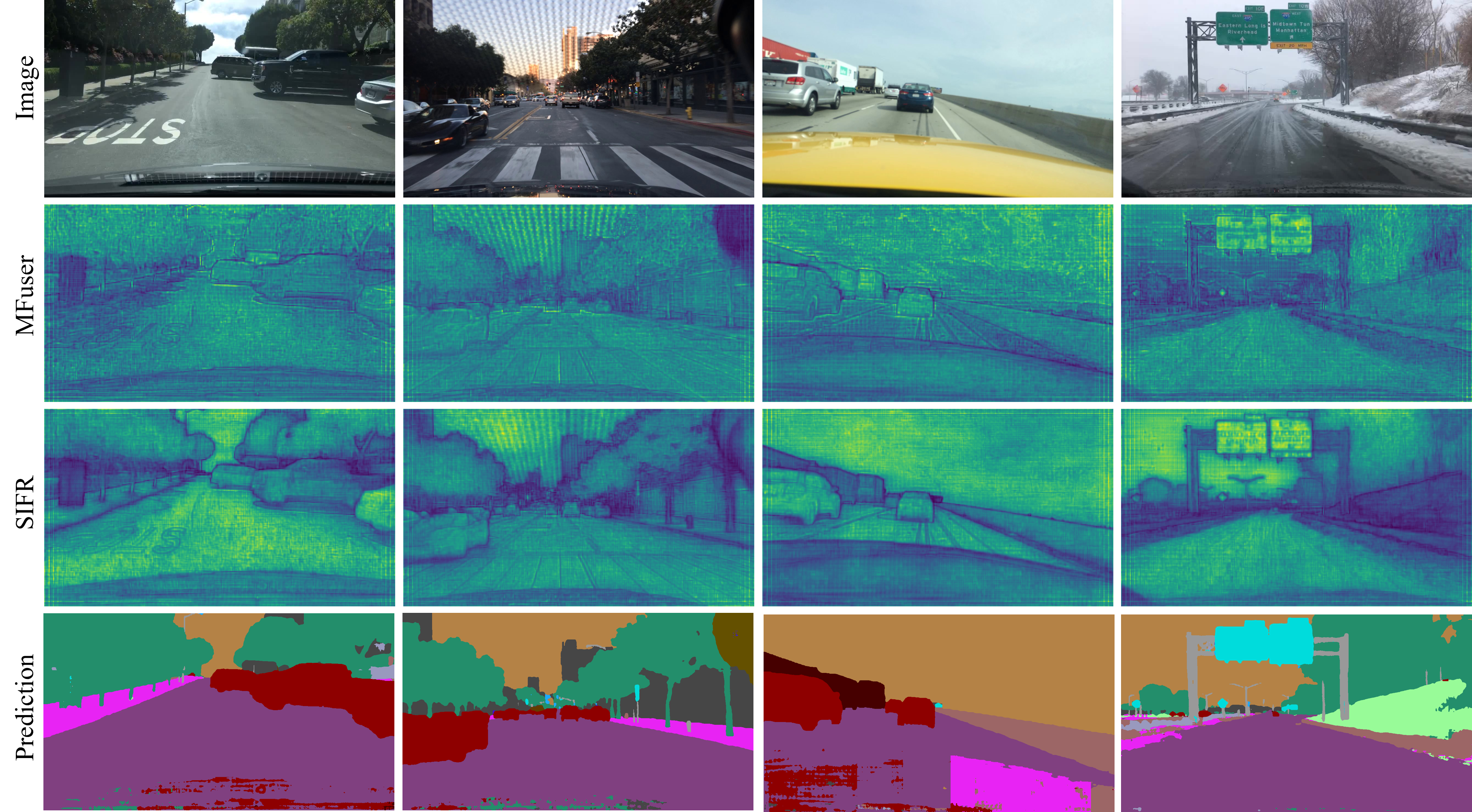}
    \caption{\textbf{Qualitative comparison of feature responses.} From top to bottom are the input images, the feature visualizations of MFuser~\cite{zhang2025mamba}, the feature visualizations of our SIFR, and the corresponding predictions of SIFR. SIFR exhibits clearer structural patterns and more discriminative feature responses than MFuser~\cite{zhang2025mamba}, particularly in regions containing distant objects and ambiguous semantic boundaries.}
    \label{fig:feature}
\end{figure*}

\subsection{Ablation  Analysis}

We conduct ablation experiments under the \textit{Cityscapes~\cite{cordts2016cityscapes} $\rightarrow$ BDD100k~\cite{yu2020bdd100k} + Mapillary~\cite{neuhold2017mapillary}} setting to analyze the effect of the proposed spectral designs from five aspects, including normalization strategy, pooling design,  module composition,
\textcolor{blue}{the number of learnable query
tokens in SGIF, and the low-pass filter ratio in DS-Norm.}

\noindent\textbf{Effect of Normalization Strategy.}
Tab.~\ref{tab:ablation_norm} compares different normalization strategies in the decoder refinement stage. 
Among all variants, the proposed DS-Norm achieves the best average performance, improving the average mIoU to 73.04.
Compared with conventional normalization schemes, such as LayerNorm, BatchNorm, GroupNorm, and InstanceNorm, DS-Norm yields more consistent gains on both BDD100k~\cite{yu2020bdd100k} and Mapillary~\cite{neuhold2017mapillary}.
This result indicates that directly applying generic normalization operators is insufficient to address domain-specific distribution shifts in decoder features. 
By contrast, our DS-Norm explicitly normalizes features with low-frequency statistics, which better suppresses residual domain-sensitive deviations and produces more robust representations for cross-domain segmentation.

\noindent\textbf{Effect of Pooling Design.}
Tab.~\ref{tab:ablation_pooling} studies different pooling combinations for generating the spectral mask in the decoder stage. Using only two types of pooled descriptors already brings reasonable performance, but the full combination of min pooling, average pooling, and max pooling gives the best result, reaching 73.04 average mIoU.
This suggests that different pooling operators capture complementary statistics in the frequency domain.
In particular, max pooling emphasizes strong spectral responses, average pooling captures global frequency trends, and min pooling further supplements suppressed yet potentially informative cues.
Their combination therefore enables a more comprehensive spectral characterization and leads to stronger domain-generalized segmentation performance.

\noindent\textbf{Effect of SGIF and SDR.}
Tab.~\ref{tab:ablation_FFT} further evaluates the contributions of the two proposed modules.
Introducing either SGIF or SDR alone already improves the cross-domain performance over the baseline, indicating that both encoder-side spectral fusion and decoder-side spectral refinement are individually effective.
Using both modules together yields further gains, highlighting their complementarity.
Specifically, SGIF strengthens intermediate fused representations by combining token interaction with frequency-guided invariant cues, while SDR removes residual domain-sensitive bias during decoding.
These results show that spectral modeling at both the feature fusion and decoder refinement stages is important for stronger robustness under domain shift.

\noindent\textbf{Hyperparameter Sensitivity.}
Table~\ref{tab:hyperparameter_sensitivity} evaluates the
sensitivity of SIFR to the number of learnable query tokens
$L_q$ in SGIF and the low-pass filter ratio $\rho$ in DS-Norm.
For each experiment, we vary one parameter while keeping the
remaining configurations fixed.

For the number of query tokens, SIFR achieves average mIoU
scores of 72.49, 72.38, 73.04, and 72.04 with $L_q$ set to
32, 48, 64, and 128, respectively. The model maintains comparable
performance with moderate token numbers, while $L_q=64$ achieves
the best result. Increasing $L_q$ to 128 leads to a moderate
performance decrease, possibly because an excessive number of
tokens introduces redundant cross-prior interactions. Therefore,
we use $L_q=64$ as the default setting.

For DS-Norm, the low-pass filter ratios of 0.25, 0.50, and 0.75
yield average mIoU scores of 73.04, 72.75, and 72.94,
respectively. The maximum difference among these settings is only
0.29 mIoU, indicating that DS-Norm is relatively robust to the
size of the low-pass filter. We adopt $\rho=0.25$ because it
provides the best overall performance.
\begin{table}[!t]
    \centering
    \setlength{\abovecaptionskip}{0.1cm}
    \setlength{\belowcaptionskip}{0.1cm}
    \setlength{\tabcolsep}{4pt}
    \caption{Ablation study of different normalization strategies on cross-domain semantic segmentation. All experiments are conducted under the \textit{Cityscapes $\rightarrow$ BDD100k + Mapillary} generalization setting.}
    \small
    \renewcommand{\arraystretch}{1.1}
    \begin{tabular}{l|cc|c}  
        \toprule
        \textbf{Normalization} & \textbf{BDD100k} & \textbf{Mapillary} & \textbf{Avg} \\
        \midrule
        Layer Norm      & 66.89 & 76.68 & 71.79 \\
        Batch Norm      & 66.09 & 77.93 & 72.01 \\
        Group Norm      & 66.56 & 77.84 & 72.20 \\
        Instance Norm   & 67.04 & 78.12 & 72.58 \\  
        \rowcolor{gray!20}
        \textbf{DS-Norm} & \textbf{67.60} & \textbf{78.48} & \textbf{73.04} \\ 
        \bottomrule
    \end{tabular}
\label{tab:ablation_norm}
\end{table}

\begin{table}[!t]
    \centering
    \setlength{\abovecaptionskip}{0.1cm}
    \setlength{\belowcaptionskip}{0.1cm}
    \setlength{\tabcolsep}{6pt}
    \renewcommand{\arraystretch}{1.1}
    \caption{Ablation study of different pooling designs under the \textit{Cityscapes $\rightarrow$ BDD100k + Mapillary} generalization setting.}
    \scriptsize
    \resizebox{0.48\textwidth}{!}{
    \begin{tabular}{ccc|cc|c}  
        \toprule
        \textbf{Min} & \textbf{Avg} & \textbf{Max} & \textbf{BDD100k} & \textbf{Mapillary} & \textbf{Avg} \\
        \midrule
        $-$ & $-$ & $-$ & 66.86 & 76.58 & 71.72\\
        $\checkmark$ & $\checkmark$ & $-$ & 66.96 & 76.68 & 71.79 \\
        $\checkmark$ & $-$ & $\checkmark$ & 67.07 & 77.93 & 72.01 \\
        $-$ & $\checkmark$ & $\checkmark$ & 66.96 & 77.84 & 72.20 \\
        \rowcolor{gray!20}
        $\checkmark$ & $\checkmark$ & $\checkmark$ &  \textbf{67.60} &  \textbf{78.48} &  \textbf{73.04} \\
        \bottomrule
    \end{tabular}
    }
    \label{tab:ablation_pooling}
\end{table}


\begin{table}[!t]
    \centering
    \setlength{\abovecaptionskip}{0.1cm}
    \setlength{\belowcaptionskip}{0.1cm}
    \setlength{\tabcolsep}{6pt}
    \caption{Ablation study of SGIF and SDR under the \textit{Cityscapes $\rightarrow$ BDD100k + Mapillary} generalization setting.}
    \tiny
    \renewcommand{\arraystretch}{1.1}
    \resizebox{0.48\textwidth}{!}{
    \begin{tabular}{cc|cc|c}
        \toprule
        \textbf{SGIF} & \textbf{SDR} & \textbf{BDD100k} & \textbf{Mapillary} & \textbf{Avg} \\
        \midrule
        $-$ & $-$  &65.60 & 76.16 & 70.88 \\
        $\checkmark$  & $-$ & 66.86 & 76.58 & 71.72 \\
        $-$ & $\checkmark$ & 65.88 & 76.53 &71.20  \\
        \rowcolor{gray!20}
        $\checkmark$ & $\checkmark$ & 67.60 & 78.48 & 73.04 \\
        \bottomrule
    \end{tabular}
    }
    \label{tab:ablation_FFT}
\end{table}

\begin{table}[t]
\centering
\caption{\textcolor{blue}{Sensitivity analysis of the number of learnable query
tokens $L_q$ in SGIF and the low-pass filter ratio $\rho$ in
DS-Norm under the Cityscapes $\rightarrow$ BDD100K + Mapillary
generalization setting.}}
\label{tab:hyperparameter_sensitivity}
\setlength{\tabcolsep}{4pt}
{\color{blue}
\begin{tabular}{lcccc}
\toprule
Hyperparameter & Value & BDD100K & Mapillary & Avg. \\
\midrule
\multirow{4}{*}{query tokens $L_q$} & 32  & 66.80 & 78.18 & 72.49 \\
                      & 48  & 66.62 & 78.13 & 72.38 \\
                      & \textbf{64}  & \textbf{67.60} & \textbf{78.48}
                      & \textbf{73.04} \\
                      & 128 & 66.18 & 77.89 & 72.04 \\
\midrule
\multirow{3}{*}{Low-pass ratio $\rho$} & \textbf{0.25} & \textbf{67.60} & \textbf{78.48}
                      & \textbf{73.04} \\
                      & 0.50 & 67.23 & 78.27 & 72.75 \\
                      & 0.75 & 67.53 & 78.34 & 72.94 \\
\bottomrule
\end{tabular}}
\end{table}

\begin{figure*}[t]
    \centering
    \vspace{-1mm}
    \includegraphics[width=\textwidth]{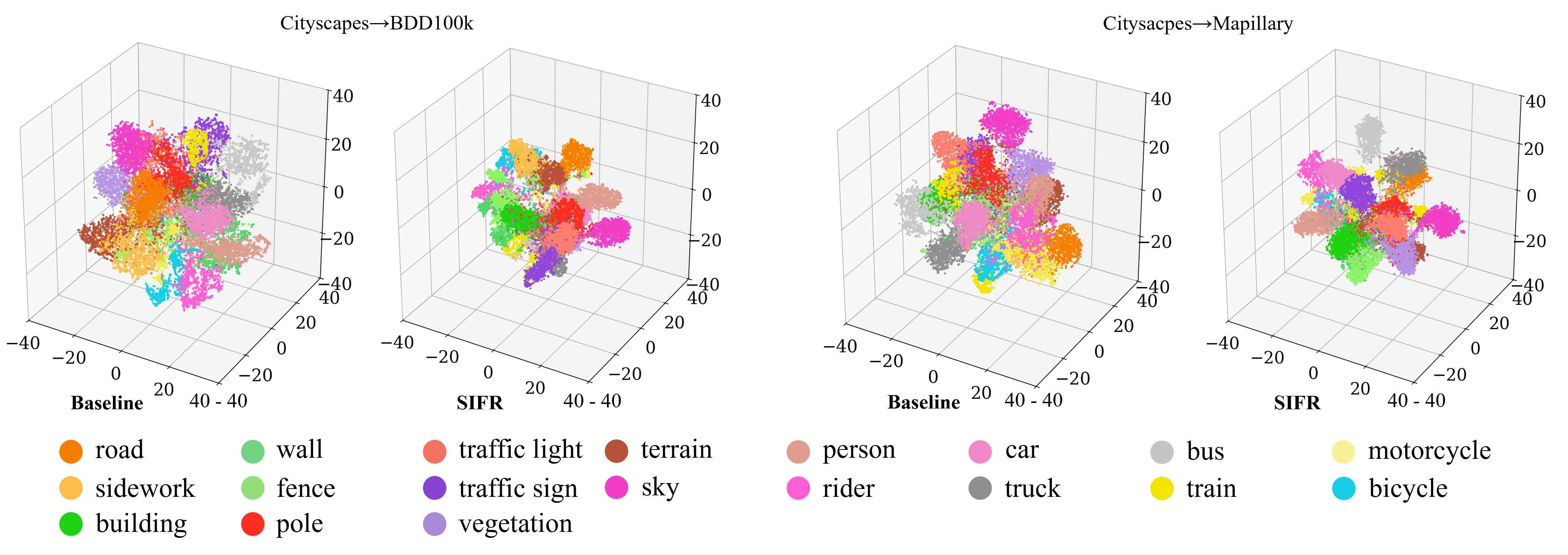}
    \vspace{-3mm}
    \caption{t-SNE visualization of decoder feature embeddings on $C\to{B,M}$. Compared with the baseline, SIFR yields more compact intra-class clusters and clearer inter-class separation across most semantic categories, indicating that it learns more discriminative and transferable representations under domain shift. The visualization is based on the default CLIP-based VLM setting.}
    \vspace{-2mm}
    \label{fig:tsne}
\end{figure*}

\begin{table}[t]
\centering
\caption{\textcolor{blue}{Quantitative representation analysis on original decoder feature embeddings. 
Lower Intra and DBI indicate more compact intra-class distributions, while higher Inter/Intra and CH indicate better inter-class separability.}}
\label{tab:representation_metrics}
\resizebox{\columnwidth}{!}{
\color{blue}
\begin{tabular}{lccccc}
\hline
Method & Domain & Intra$\downarrow$ & Inter/Intra$\uparrow$ & DBI$\downarrow$ & CH$\uparrow$ \\
\hline
Baseline & BDD100K & 4.4342 & 1.5037 & 2.4203 & 1103.83 \\
 \textbf{SIFR  }   &  \textbf{BDD100K} & \textbf{3.9008} & \textbf{2.1936} & \textbf{1.7189} & \textbf{2217.43} \\
\hline
Baseline & Mapillary & 4.7610 & 1.6438 & 1.7518 & 2561.98 \\
 \textbf{SIFR}     & \textbf{Mapillary} & \textbf{4.5881} & \textbf{1.8650} & \textbf{1.3806} & \textbf{3175.47} \\
\hline
\end{tabular}
}
\end{table}

\begin{figure}[t]
    \centering
    \includegraphics[width=0.5\textwidth]{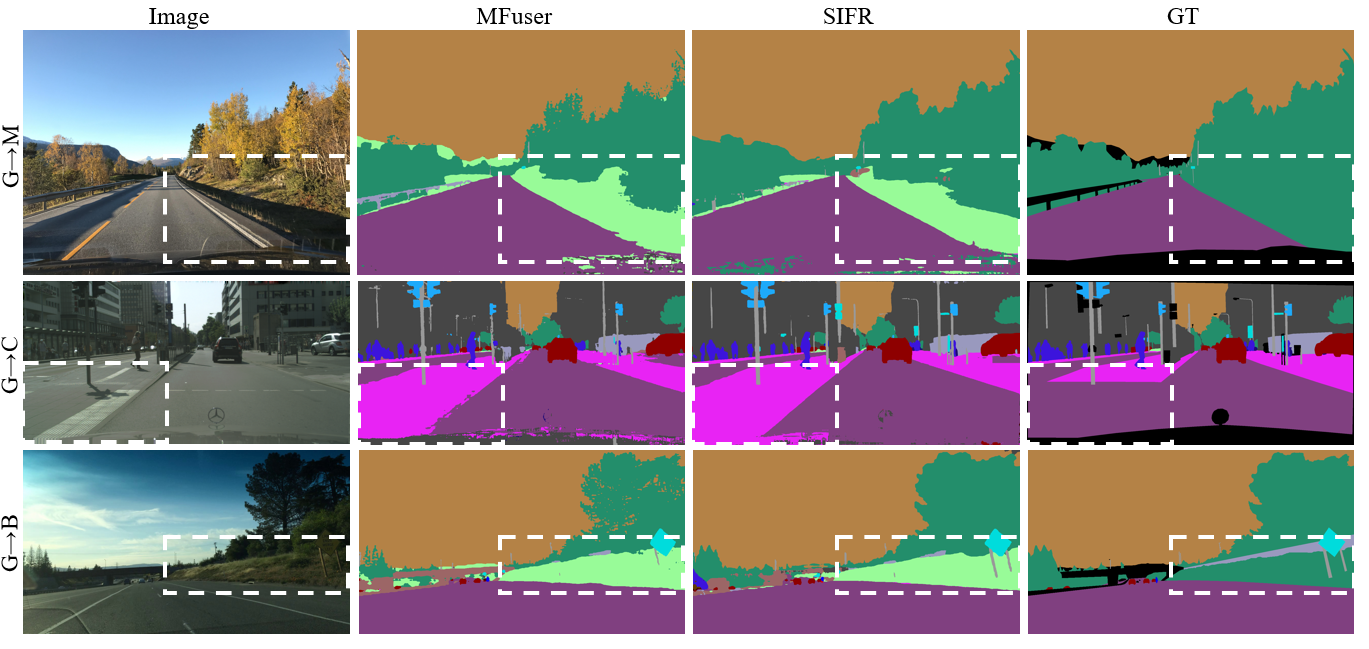}
    \caption{\textbf{Representative failure cases of the proposed method.}
From left to right, each row presents the input image, the SIFR
    prediction, and the ground-truth annotation. In the first and
    third cases, parts of the roadside vegetation are incorrectly
    classified as terrain because mixed natural elements, such as
    dry grass, rocks, shrubs, and autumn-colored trees, exhibit
    similar appearances. In the second case, parts of the drivable
    road are incorrectly classified as sidewalk because of their
    similar pavement appearances and ambiguous boundaries.
    Black regions in the ground-truth annotations denote ignored
    pixels and are excluded from the evaluation.}
    \label{fig:failure_case}
\end{figure}

\subsection{Representation Analysis}

\noindent\textbf{Local Feature Response analysis.}
We first visualize the intermediate feature responses in Fig.~\ref{fig:feature} to examine how SIFR affects local semantic and structural cues under domain shift. Compared with MFuser, SIFR produces clearer and more concentrated responses on distant objects, thin structures, and semantic boundaries. In particular, the highlighted regions show that SIFR better preserves object contours and suppresses irrelevant activations in ambiguous areas. These observations suggest that the proposed spectral designs help the model focus on structurally stable and semantically informative patterns, which are more transferable across domains.

{\color{blue}
\noindent\textbf{Global Feature Space Analysis.}
To examine the representation quality learned by the whole framework, we visualize decoder-level embeddings using t-SNE in Fig.~\ref{fig:tsne}. Specifically, we extract the 256-dimensional mask features produced by the final mask-feature projection of the pixel decoder immediately before mask prediction, using the same decoder stage for both the baseline and SIFR. For the $C \rightarrow B$ and $C \rightarrow M$ settings, features are sampled from the unseen BDD100K and Mapillary validation sets, respectively. To alleviate class imbalance, we sample up to 2,000 pixel-level feature vectors from each of the 19 shared semantic categories and project them into a three-dimensional space using t-SNE with PCA initialization and a fixed random seed, with each point colored according to its ground-truth semantic category.

Compared with the baseline, SIFR produces more compact intra-class clusters and clearer inter-class separation on both target domains. To further quantify this observation, we compute representation-level criteria on the original decoder embeddings before t-SNE dimensionality reduction, including intra-class feature distance, inter-/intra-class distance ratio, Davies-Bouldin Index (DBI), and Calinski-Harabasz Index (CH). Lower intra-class distance and DBI indicate more compact
class-wise feature distributions, whereas higher inter-/intra-class
distance ratio and CH indicate better inter-class separability. As shown in Table~\ref{tab:representation_metrics}, SIFR consistently improves these metrics on both BDD100K and Mapillary. On BDD100K, SIFR reduces the intra-class distance from 4.4342 to 3.9008 and DBI from 2.4203 to 1.7189, while increasing the inter-/intra-class ratio from 1.5037 to 2.1936 and CH from 1103.83 to 2217.43. On Mapillary, SIFR also achieves lower intra-class distance and DBI, together with higher inter-/intra-class ratio and CH. These results provide
quantitative evidence that SIFR learns more compact and discriminative
representations on unseen domains, thereby supporting its improved
cross-domain generalization.
}

{\color{blue}
\noindent\textbf{Failure Case Analysis.}
Although the proposed method achieves favorable generalization performance
across different target domains, it still encounters difficulties in
distinguishing visually similar categories and preserving fine-grained
semantic boundaries, as illustrated in Fig.~\ref{fig:failure_case}. A recurring failure pattern
observed in the first and third examples is the confusion between vegetation
and terrain. In these cases, parts of the roadside vegetation are incorrectly
classified as terrain because mixed natural elements, such as dry grass,
rocks, shrubs, and irregular surfaces, exhibit similar colors and textures.
In the second example, parts of the drivable road are incorrectly classified
as sidewalk, particularly in regions where the two categories have similar
pavement appearances and lack clear curb boundaries. Moreover, the boundaries
of some small and thin structures are not accurately preserved.}


\section{Conclusion}
In this paper, we presented SIFR, a frequency-aware dual-stream framework for domain generalized semantic segmentation that leverages the complementary strengths of frozen CLIP and DINOv2 under domain shift. SIFR improves generalization through both heterogeneous prior fusion and decoder refinement. It introduces Spectral-Guided Invariant Fusion to enhance semantic–structural interaction while reducing domain-sensitive interference, and Spectral Decoder Refinement to stabilize multi-scale decoding using low-frequency statistics-based normalization and adaptive spectral enhancement. Experiments on synthetic-to-real, real-to-real, and real-to-adverse-weather  benchmarks show that SIFR consistently improves cross-domain segmentation, highlighting frequency-aware modeling as an effective way to exploit heterogeneous foundation priors for DGSS.


\bibliographystyle{IEEEtran}
\bibliography{refs}

\end{document}